# An Agentic AI Workflow for Detecting Cognitive Concerns in Real-world Data

Jiazi Tian, MSc[1] ; Liqin Wang, PhD[2] ; Pedram Fard, PhD[3] ; Valdery Moura Junior, PhD, MBA[1] ; Deborah Blacker, MD, ScD[4] ; Jennifer S. Haas, MD, MSc[1] ; Chirag Patel, PhD[3] ; Shawn N. Murphy, MD, PhD[5] ; Lidia M.V.R. Moura, MD, PhD, MPH[5]*; Hossein Estiri, PhD[1]*

[1] Department of Medicine, Massachusetts General Hospital, Boston, MA, USA
[2] Department of Medicine, Brigham and Women's Hospital, Boston, MA, USA
[3] Department of Biomedical Informatics, Harvard Medical School, Boston, MA, USA
[4] Department of Psychiatry, Massachusetts General Hospital, Boston, MA, USA
[5] Department of Neurology, Massachusetts General Hospital, Boston, MA, USA
* contributed equally

Early identification of cognitive concerns is critical but often hindered by subtle symptom presentation. This study developed and validated a fully automated, multi-agent AI workflow using LLaMA 3 8B to identify cognitive concerns in 3,338 clinical notes from Mass General Brigam. The agentic workflow, leveraging task-specific agents that dynamically collaborate to extract meaningful insights from clinical notes, was compared to an expert-driven benchmark. Both workflows achieved high classification performance, with F1-scores of 0.90 and 0.91, respectively. The agentic workflow demonstrated improved specificity (1.00) and achieved prompt refinement in fewer iterations. Although both workflows showed reduced performance on validation data, the agentic workflow maintained perfect specificity. These findings highlight the potential of fully automated multi-agent AI workflows to achieve expert-level accuracy with greater efficiency, offering a scalable and cost-effective solution for detecting cognitive concerns in clinical settings.

# Introduction

Recent advancements in large language models (LLMs) have significantly enhanced natural language processing and artificial intelligence, demonstrating impressive capabilities in language generation and contextual understanding.[1–4] These innovations promise to improve clinical workflows by efficiently processing and interpreting complex medical data.[5,6]

Integrating LLMs into dementia screening offers transformative potential for early detection and intervention.[7,8] Early identification of cognitive decline is crucial for effective treatments, such as FDA-approved beta-amyloid-targeting drugs like lecanemab and aducanumab.[7,9,10] However, traditional screening methods like the Mini-Mental State Examination (MMSE) and Montreal Cognitive Assessment (MoCA) often face challenges in accessibility and scalability due to their reliance on in-person administration.[11–13]

Emerging research shows that LLMs can address these limitations by providing scalable, cost-effective screening tools. They can analyze language patterns - such as syntax, semantics, and lexical diversity - to detect subtle cognitive impairments indicative of early-stage dementia with high accuracy.[14,15] Additionally, LLMs can be integrated into telehealth platforms, enabling remote screening and increasing access for underserved populations.[16,17]

Prompt engineering is crucial for optimizing LLM performance, especially in high-precision clinical applications where subtle phrasing changes can cause inconsistent or inaccurate results.[18,19] However, manually refining prompts is time-consuming and requires specialized expertise. To overcome these challenges, recent research has shown that LLMs can power autonomous agents capable of mimicking human behavior in interactive, context-sensitive tasks. This advancement enables more efficient and reliable applications, such as detecting cognitive concerns in clinical notes, without the extensive trial-and-error typically needed for prompt engineering.[20,21]

In this study, we examined the accuracy of a novel agentic AI workflow for screening patient charts to identify evidence of cognitive concerns. The workflow utilizes specialized AI agents assigned to specific subtasks within the classification process, leveraging the reasonably accessible LLaMA 3 8B.[22]

# Methodology

The primary aim of this study was to develop and test an automated multi-agent AI workflow that can achieve high classification accuracy in detecting cognitive concerns while balancing sensitivity and specificity.

## Clinical Setting and Data Sources

We collected clinical notes documented between January 1, 2016, and December 31, 2018, from Mass General Brigham (MGB)'s Research Patient Data Registry (RPDR), including history and physical exam notes, clinic visit notes, discharge summaries, and progress notes. Instead of segmenting the notes into chunks or sections, we utilized the full context of the original notes in this study. The data was analyzed between February 1, 2024, and August 31, 2024.

## Creation of Datasets

The study cohort included 3,338 clinical notes from 200 Mass General Brigham patients who were classified into two categories—patients with cognitive concerns and those without—based on manual chart review.[23] These classifications were used as the reference standard for this study. We framed the identification of cognitive concerns as a binary classification task. We split this data randomly to curate a prompt refinement data set and an out-of-sample validation data set to evaluate the performance and generalizability of the agentic workflow and a benchmark expert-driven workflow.

## Large Language Model

We utilized the LLaMA 3 8B[22], developed by Meta AI, which is an open-source LLM capable of running locally, thereby ensuring the privacy of sensitive data. We obtained access to the model's weights through Meta and the Hugging Face library. We configured the temperature parameter to 0.1 to allow some flexibility in answer generation while mitigating potential noise from the lengthy original notes. The maximum output token length was set to 256, with all other parameters maintained at their default settings. The model was deployed on a local Linux-based server with 48 Cores (96 Threads) and 256GB of high-speed RAM.

## Study Design

As illustrated in Figure 1, two parallel workflows were developed. An agentic workflow aimed at classifying patient notes in an automated process, in which specialized agents made inferences through communications and using tools. An expert-driven workflow was created as a benchmark. Initially, the model operated using a zero-shot approach to determine whether the notes indicated cognitive concerns. Following this, generated knowledge prompting[24] was applied to improve the performance of the model's responses. The prompt configuration included both system and user prompts: the system prompt defined the LLM's role, and the user prompt provided task-specific information and questions.

Since each patient had at least one clinical note and the LLM operated on individual notes, a patient's cognitive status was determined by aggregating and analyzing all the responses generated by the LLM for their respective notes. Specifically, if any of the responses indicated the presence of cognitive concerns, the patient was labeled as "with cognitive concerns."

Conversely, if all responses consistently indicated an absence of cognitive concerns, the patient was labeled as "without cognitive concerns." In cases where the outputs were misformatted (e.g., not adhering to the required binary "yes/no" response), non-informative or inconclusive, the patient was assigned the label "uncertain," which was excluded from further analysis.

The initial prompt (P0) to begin both workflows was "Is this note indicative of any cognitive concern, yes or no?" (Table 2).

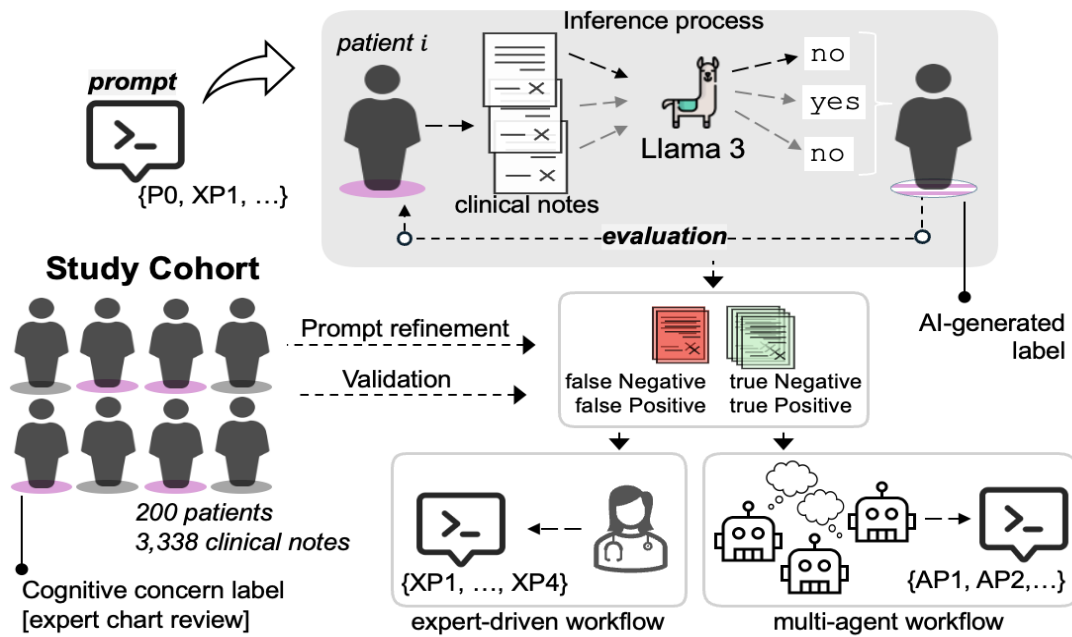

**Figure 1. Study Flow Diagram.** The study cohort is split into a prompt refinement and a validation set (for evaluating generalizability). The process begins with prompt 0 (P0: `Is this note indicative of any cognitive concern, yes or no? \n {note}`). AI-generated labels are evaluated against chart-reviewed labels (i.e., ground truth). An expert-driven approach is developed as a benchmark to compare with the performance of the fully-automated agentic workflow. The multi-agent – i.e., agentic – workflow suggests new prompts by evaluating cases of misclassification and summarization of suggestions from specialized agents. Abbreviations: P0: initial prompt; XP1, XP4: expert prompt 1, expert prompt 4; AP1, AP2: agent prompt 1, agent prompt 2.

## Agentic Workflow

The agentic workflow comprised six specialized agents to automatically refine the prompt and enhance the capabilities of LLM for cognitive concern detection.

**Specialist**: the role is an expert in evaluating patients with cognitive concerns. It takes the prompt and clinical notes as input and generates the corresponding responses, providing a "yes" or "no" answer along with the reason behind the decision.

**Evaluator**: the role is a helpful assistant. It determines the label through the aggregation of the Specialist's responses across all clinical notes of that patient. It then compares these labels to the reference labels and generates an evaluation metrics report, including sensitivity, specificity, negative predictive value (NPV), positive predictive value (PPV), accuracy, and F1-score.

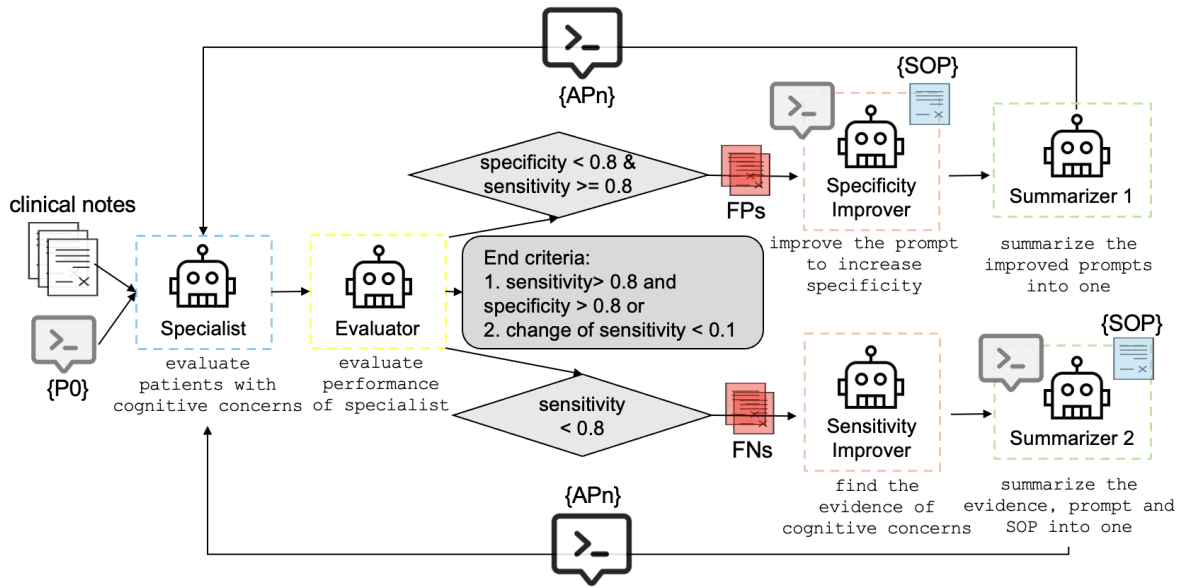

**Figure 2. The agentic workflow, including six specialized agents.** Abbreviations: P0: initial prompt; APn: agent prompt number n; FPs: false positive cases; FNs: false negative cases; SOP: standard operating procedure.

**Specificity improver:** the role is that it has expertise in both clinical knowledge and advanced prompt engineering technologies. The standard operating procedure (SOP), developed in the previous study[23], is given as the guidelines for identifying cognitive concerns. It analyzes false positive examples to identify reasons for misclassification by the LLM and uses this understanding to refine the Specialist's prompt to increase the specificity.

**Sensitivity improver**: the role is that it has expertise in both clinical knowledge and advanced prompt engineering technologies. It focuses on false negative cases, seeking evidence of cognitive concerns that the LLM may have missed, thereby improving sensitivity.

**Summarizer 1**: the role is a helpful assistant. It summarizes the improvements suggested by the Specificity improver for each case into one new improved prompt, which will be sent back to the Specialist.

**Summarizer 2**: the role is that it has expertise in both clinical knowledge and advanced prompt engineering technologies. It first summarizes the findings from the Sensitivity improver. It then refines the prompt by incorporating both the findings and the SOP to enhance performance. This improved prompt will be sent back to the Specialist.

As shown in Figure 2, if the sensitivity falls below 0.8, the workflow directs the task to the Sensitivity Improver, followed by Summarizer 1. The improved prompt is then provided to the Specialist for another iteration. A similar process is followed for specificity enhancement. Through this iterative process, the agents collaboratively refine the LLM's ability to detect cognitive concerns, continuously optimizing performance by improving sensitivity and specificity with targeted adjustments and evaluations. The workflow is configured with a default maximum of three iterations. The process concludes when either both sensitivity and specificity achieve a threshold of at least 0.8, or when the change in sensitivity falls below 0.1.

## Expert-Driven Workflow

The expert-driven workflow combined the capabilities of the LLM with the clinical expertise, who is an experienced Neurologist. As shown in Figure 3, the process began with the initial prompt to generate responses for individual notes. Labels were then assigned to each patient by aggregating the responses across all their clinical notes. Next, we generated an evaluation metrics report by comparing these aggregated patient-level labels to the corresponding reference labels. False positive and false negative cases were then randomly chosen for review by the clinical expert. Based on the clinician's feedback, the prompt configurations were revised, including both system and user contents. The improved prompt was resubmitted to the LLM for new testing. To further optimize the prompt formulations, we utilized ChatGPT-4o through the user interface.

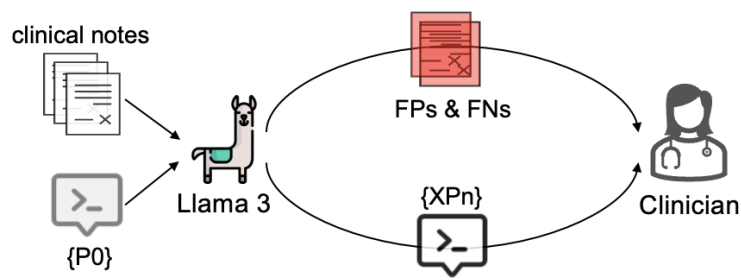

**Figure 3. The Expert-Driven workflow.** Abbreviations: P0: initial prompt; XPn: expert prompt number n; FPs: false positive cases; FNs: false negative cases.

## Evaluation

We conducted a comprehensive performance evaluation on the two workflows by calculating sensitivity, specificity, accuracy, PPV, NPV, and F1-score, against the chart-reviewed (ground truth) labels. Since the patient-level labels included three categories, we excluded the 'uncertain' cases from the calculation of evaluation metrics. Furthermore, to evaluate the generalizability of the two workflows, we applied all prompt configurations to the independent validation dataset and evaluated their ability to accurately detect cognitive concerns from clinical notes.

# Results

In total, 3,338 clinical notes from 200 patients were included in the study. Table 1 shows the characteristics of the two datasets. Patients in the prompt refinement dataset (2,228 notes) had an average age of 77.9 years and were 60 percent female. 50 percent of patients in this dataset had validated indications of cognitive concerns. The independent validation dataset (1,110 notes), included patients with an average age of 76.1 years, 59 percent of whom were female. Indications of cognitive concerns were present in 33% of patients in the validation data set. The majority of the population in both datasets was White and non-Hispanic.

Table 1. Summary statistics of the study population.

|  | Prompt refinement | Validation |
|---|---|---|
| Number of patients | 100 | 100 |
| Number of notes | 2,228 | 1,110 |
| Age, mean (SD) years | 77.9 (7.07) | 76.1 (7.02) |
| Sex, n (%) |  |  |
|   Female | 60 (60.0%) | 59 (59.0%) |
|   Male | 40 (40.0%) | 41 (41.0%) |
| Race |  |  |
|   White | > 85 (>85%) | > 85 (>85%) |
|   Black | < 10 (<10%) | < 10 (<10%) |
|   Asian | < 10 (<10%) | < 10 (<10%) |
|   Others or Unknown | < 10 (<10%) | < 10 (<10%) |
| Ethnic group |  |  |
|   Non-Hispanic | > 85 (>85%) | > 85 (>85%) |
|   Unknown | < 10 (<10%) | < 10 (<10%) |
| Presence of cognitive concerns, n (%) | 50 (50.0%) | 33 (33.0%) |

Abbreviation: SD: standard deviation.

## Agentic Workflow

A total of two iterations were needed for the agentic workflow to achieve the minimum thresholds for sensitivity and specificity (0.8). The refined prompts (AP1, AP2) from each iteration are presented in Table 2.

**Table 2. Prompts generated throughout the agentic workflow.**

| Versions of Prompt | Prompt |
|---|---|
| P0 | *System*: You are a neurologist.<br><br>*User*: Is this note indicative of any cognitive concern, yes or no? \n {note} |
| AP1 | Is this note indicative of any cognitive concern, yes or no? Please consider the patient's medical history, symptoms, and physical examination findings, as well as any relevant keywords or phrases, to determine if there are any red flags for cognitive impairment or decline. |
| AP2 | Is this note indicative of any cognitive concern, yes or no?<br><br>This question can be answered by reviewing the note for mentions of:<br><br>  Medications commonly used to treat cognitive impairment (e.g. Donepezil, Rivastigmine, Galantamine, Memantine)<br><br>  Diagnoses or symptoms related to dementia, Alzheimer\'s disease, or memory issues<br><br>  Cognitive assessments (e.g. MOCA, MMSE)<br><br>  Behavioral symptoms (e.g. agitation, aggression, delirium, hallucinations)<br><br>  Reported cognitive changes or objective evidence of cognitive impairment<br><br>  Preserved independence in functional abilities<br><br>  Absence of delirium or major psychiatric disorders<br><br>If any of these criteria are met, the answer is "yes". Otherwise, the answer is "no". |

Abbreviations: P0: initial prompt; AP: agent prompt.

During the first run, the SOP was provided to Summarizer 1 instead of the Specificity Improver agent, which resulted in outcomes contrary to our expectations, with no improvement in specificity, which decreased from 0.2 to 0.05 compared with P0, as shown in Table 3. In the second iteration, the SOP was provided to the Specificity Improver agent, leading to the expected increase in specificity from 0.20 to 1.00, while prompt AP2 classified 33 cases as uncertain (eTable 2). The overall performance of AP2 was more balanced as the F1-score was increased from 0.70 to 0.91. Although sensitivity decreased from 0.94 to 0.84 and NPV did not change a lot from 0.75 to 0.77, there were notable improvements in accuracy and PPV, which increased from 0.59 and 0.56 to 0.90 and 1.00, respectively.

Table 3. Classification performance of all prompts from agentic workflow on prompt refinement data set.

| Performance | Sensitivity | Specificity | PPV | NPV | Accuracy | F1-score |
|---|---|---|---|---|---|---|
| P0 | 0.94 | 0.2 | 0.56 | 0.75 | 0.59 | 0.7 |
| AP1 | 1 | 0.05 | 0.64 | 0.75 | 0.55 | 0.7 |
| AP2 | 0.84 | 1 | 1 | 0.77 | 0.9 | **0.91** |

Abbreviations: P0: initial prompt; AP: agent prompt; PPV: positive predictive value; NPV: negative predictive value
* NPV was not calculable because there were no true or false negative cases.

## Expert-driven workflow

As presented in eTable 1, clinicians generated four distinct prompts (XP1, XP2, XP3 and XP4) in the expert-driven workflow to refine the model's performance. After the clinician reviewed a randomly selected subset of false positive and false negative cases from P0, the scope of cognitive concerns was added to the subsequent prompts. Furthermore, the model's precise role was elucidated. These modifications were incorporated into the XP1 and XP2 prompts. Besides, we refined the XP2 by adding one more question.

The XP3 prompt was created to exclude risk factors or normal screening results from being used as evidence of cognitive concerns and to avoid making assumptions based on limited information. Lastly, XP4 was refined using ChatGPT-4o through the user interface, with the input: *"We found that prompt 3 works better than prompt 1 and prompt 2. However, using prompt 3, the model is likely to use the risk factors or standard screening with normal results as evidence of cognitive concerns to make the assumptions. In this case, how to improve the prompt 3?"*.

The performance of the prompts from the expert-driven workflow is shown in eTable 2. A comparison of the results from prompts P0, XP1, and XP2 revealed varying levels of performance improvement. The specificity of XP1 increased to 0.32, and the F1-score increased to 0.76, though sensitivity slightly decreased to 0.92, and the number of uncertain cases rose to 14. Prompt XP2 demonstrated extreme sensitivity and specificity, 1.0 and 0.0 respectively, by classifying all cases as with cognitive concerns but two as uncertain, with a slight reduction in the F1-score to 0.67.

Prompts XP3 and XP4 classified more cases as uncertain, 35 and 21, respectively, compared to the results of previous prompts (see eTable 3). XP3 reached a high F1-score of 0.86, a high specificity of 1.00, and a sensitivity of 0.76. Overall, XP4 outperformed all other prompts, with an F1-score of 0.90, demonstrating a strong balance between sensitivity (0.87) and specificity (0.91) in detecting cognitive concerns. The model also showed high precision, with a PPV of 0.93 and an NPV of 0.93, meaning it accurately identified both positive and negative cases. The overall accuracy of the model was 0.89, indicating that 89% of classifications matched the reference labels.

## Evaluating generalizability

To evaluate the generalizability of the agentic workflow, we computed performance metrics for the two workflows based on the ground truth labels in the validation set (eTable 4). The final prompt from the expert-driven workflow (XP4) demonstrated the most balanced performance, achieving an F1-score of 0.79, a sensitivity of 0.70, a specificity of 0.97, a PPV of 0.92, an NPV of 0.86 and an accuracy of 0.88 with only 2 uncertain cases (eTable 5). The AP2 achieved an F1-score of 0.76, a sensitivity of 0.61, and a specificity of 1.00, with 2 uncertain cases.

Consistent with the performance on the prompt refinement set, in the expert-driven workflow, prompt XP4 demonstrated the most balanced performance on the validation set. The agentic workflow demonstrated an improvement over the expert-driven workflow on the prompt refinement set, however, when applied to the validation set, the agentic workflow did not show any significant enhancement in performance compared to the expert-driven approach.

# Discussion

In this study, we developed and evaluated a novel, fully automated agentic AI workflow that leverages specialized AI agents for discrete tasks, utilizing the computationally efficient LLaMA 3 8B to screen patient charts for indicators of cognitive concerns. This multi-agent approach was assessed against an expert-driven benchmark workflow, which refines prompts through iterative review. Our findings demonstrate that LLaMA 3 8B effectively identifies cognitive concern indicators in clinical notes, highlighting its potential as a moderately accessible tool for clinical screening. Notably, our results suggest that a fully automated agentic workflow can achieve comparable performance to expert-driven methods while requiring fewer iterations (2 vs. 4), thereby reducing the associated resource burden. While challenges such as overfitting remain, these limitations may be mitigated by further specialization of agents. Overall, agentic

AI workflows present a scalable and efficient solution for the automated screening of cognitive concerns across diverse clinical documentation.

Relying only on unstructured clinical notes is insufficient for dementia research and clinical tool development, as it requires resource-intensive efforts to integrate additional contextual data like medications and diagnoses for comprehensive analysis. The future of such endeavors is likely to depend on AI tools that can leverage multi-modal data sources.[25]

Despite obtaining acceptable results in this study, we encountered specific limitations and challenges. For instance, there were occasions where the LLM generated responses that deviated from the expected binary "yes" or "no" format. Examples of this included responses starting with phrases like "Based on the provided note/medical record..." or "this note is not a medical note...". Additionally, some outputs contained irrelevant instructions, such as "Please use the Print Group Designer activity...". Other non-conforming outputs included summaries of medical notes, random numerical sequences, etc..

The sensitivity of LLMs was also limited by the nature of the source data - EHRs, as cognitive concerns are not always self-reported by patients prior to a formal dementia diagnosis. They are more commonly observed and reported by caregivers or family members during routine primary care visits or noted incidentally during specialized care, such as physical therapy sessions - observations that are usually under-documented and under-represented in EHRs.[23,26]

Learning from false-negative cases was challenging due to patient-level chart reviews, making it unclear which specific notes contained cognitive concerns. Attempts to summarize these notes didn't improve sensitivity. To address this limitation, incorporating a cohort-level bias detection agent may enhance the generalizability of agentic workflows.

We employed LLaMA 3 8B, an open-source LLM with the fewest publicly available parameters. As outlined in the methodology, our rationale was to leverage a more computationally efficient model that could be feasibly implemented in low-resource settings. Future research should investigate whether utilizing more advanced models with a larger number of parameters could yield improved results.

LLMs offer a promising avenue for developing scalable and cost-effective workflows to enhance clinical care through timely assessments and interventions. Furthermore, LLMs can augment research efforts by generating comprehensive datasets that capture the earliest EHR indicators of cognitive impairment. Early identification through LLMs has the potential to optimize clinical trial recruitment by identifying candidates at the initial stages of the disease continuum, where therapeutic interventions may have the greatest efficacy.

Considering the substantial costs and resource demands associated with expert-driven workflows, our findings demonstrate that a fully automated, agentic approach can achieve comparable performance in identifying cognitive concerns while requiring fewer iterations. This efficiency underscores the potential for AI-driven agents to streamline clinical workflows, reducing the burden on healthcare professionals while maintaining diagnostic accuracy. While challenges such as overfitting remain a consideration, these limitations can be mitigated through the development of more specialized agents with enhanced domain adaptation and robustness. Ultimately, agentic workflows represent a scalable and adaptable solution for systematically screening cognitive concerns across diverse clinical notes, offering a pathway toward broader implementation in real-world healthcare settings.

## Acknowledgments

**Contribution:** H.E., L.M.M., and J.T. conceived, designed, and planned this study. H.E. and L.M.M collected and acquired the data. J.T. developed the workflow. H.E., L.M.M., and J.T. analyzed the data. H.E., L.M.M., J.T., and L.W. interpreted the data. H.E., L.M.M., J.T., and L.W. drafted the paper. All authors critically reviewed the paper. H.E., L.M.M., J.T., and L.W. revised the final paper. H.E., L.M.M., and J.T. had access to all the data in the study. All authors approved the decision to submit for publication.

**Declaration of Interests:** The authors declare no competing interests.

**Ethics approval:** Use of patient data in this study was approved by the Mass General Brigham Institutional Review Board (protocol 2020P001063).

**Funding:** This study has been supported by grants from the National Institutes of Health: the National Institute on Aging (RF1AG074372) and the National Institute of Allergy and Infectious Diseases (R01AI165535).

# Supplementary

**eTable 1**. Prompts generated throughout the expert-driven workflow.

| Versions of Prompt | Prompt |
|---|---|
| XP1 | *System*: You are an expert in reviewing patient's electronic health records and classifying these records based on current evidence of cognitive concerns.<br><br>*User*: Assume 'cognitive concern' refers to any signs of cognitive decline (memory, attention, language, visuospatial skills, executive functioning) beyond normal aging. These concerns can stem from causes like dementia, brain injuries, metabolic disorders, psychiatric conditions, medications, or substance abuse. Is this note indicative of any cognitive concern, yes or no? \n {note} |
| XP2 | *System*: You are an expert in evaluating patients with cognitive concerns.<br><br>*User*: Assume 'cognitive concern' refers to any signs of cognitive decline (memory, attention, language, visuospatial skills, executive functioning) beyond normal aging. These concerns can stem from causes like dementia, brain injuries, metabolic disorders, psychiatric conditions, medications, or substance abuse. Is this note indicative of any cognitive concern, yes or no? If yes, classify the identified cognitive concern, current/present or possible/potential? \n {note} |
| XP3 | *System*: You are an expert in evaluating patients with cognitive concerns.<br><br>*User*: Assume 'cognitive concern' refers to any signs of cognitive decline (memory, attention, language, visuospatial skills, executive functioning) beyond normal aging. These concerns can stem from causes like dementia, brain injuries, metabolic disorders, psychiatric conditions, medications, or substance abuse. Refrain from using risk factors or standard screening with normal results as evidence of cognitive concerns. Don't make assumptions with limited information. Based on this definition, is the following note indicative of any cognitive concern, yes or no? \n {note} |
| XP4 | *System*: You are an expert in evaluating patients with cognitive concerns.<br><br>*User*: Assume 'cognitive concern' refers to any signs of cognitive decline (memory, attention, language, visuospatial skills, executive functioning) beyond normal aging. These concerns can stem from causes like dementia, brain injuries, metabolic disorders, psychiatric conditions, medications, or substance abuse. Do not consider risk factors or normal screening results as evidence of cognitive concerns. Only documented symptoms, behaviors, or clinical findings indicative of cognitive decline should be considered. Is this note indicative of any cognitive concern, yes or no? \n{note} |

Abbreviations: XP: expert prompt.

An Agentic AI Workflow for Detecting Cognitive Concerns in Real-world DataeTable 2. Classification performance of all prompts from the expert-driven on prompt refinement dataset.

| Performance | Sensitivity | Specificity | PPV | NPV | Accuracy | F1-score |
|---|---|---|---|---|---|---|
| XP1 | 0.92 | 0.32 | 0.64 | 0.75 | 0.66 | 0.76 |
| XP2 | 1 | 0 | 0.5 | nan* | 0.5 | 0.67 |
| XP3 | 0.76 | 1 | 1 | 0.7 | 0.85 | 0.86 |
| XP4 | 0.87 | 0.91 | 0.93 | 0.93 | 0.89 | 0.9 |

Abbreviations: XP: expert prompt; PPV: positive predictive value; NPV: negative predictive value
* NPV was not calculable because there were no true or false negative cases.

eTable 3. Classification results from LLaMA 3 8B for prompt refinement dataset.

|  | Expert-driven workflow | | | | | Agentic workflow | |
|---|---|---|---|---|---|---|---|
|  | P0 | XP1 | XP2 | XP3 | XP4 | AP1 | AP2 |
| TN | 9 | 12 | 0 | 23 | 29 | 2 | 23 |
| FP | 36 | 25 | 49 | 0 | 3 | 42 | 0 |
| FN | 3 | 4 | 0 | 10 | 6 | 0 | 7 |
| TP | 46 | 45 | 49 | 32 | 41 | 49 | 37 |
| Uncertain | 6 | 14 | 2 | 35 | 21 | 7 | 33 |

Abbreviations: TN: true negative; FP: false positive; FN: false negative; TP: true positive

eTable 4. Classification performance of all prompts on the validation dataset.

| Performance | Expert-driven workflow | | | | | Agentic workflow | |
|---|---|---|---|---|---|---|---|
|  | P0 | XP1 | XP2 | XP3 | XP4 | AP1 | AP2 |
| Sensitivity | 0.91 | 0.82 | 1.00 | 0.58 | 0.7 | 1.00 | 0.61 |
| Specificity | 0.56 | 0.78 | 0.00 | 0.98 | 0.97 | 0.00 | 1.00 |
| PPV | 0.51 | 0.68 | 0.42 | 0.95 | 0.92 | 0.56 | 1.00 |
| NPV | 0.93 | 0.88 | nan* | 0.82 | 0.86 | nan* | 0.85 |
| Accuracy | 0.68 | 0.79 | 0.42 | 0.85 | 0.88 | 0.56 | 0.88 |
| F1-score | 0.65 | 0.74 | 0.60 | 0.72 | 0.79 | 0.72 | 0.76 |

Abbreviations: PPV: positive predictive value; NPV: negative predictive value
* NPV is not calculable because there are no true or false negative cases.

**eTable 5**. Classification results from LLaMA 3 8B for validation dataset.

| | Expert-driven workflow | | | | | Agentic workflow | |
|---|---|---|---|---|---|---|---|
| | P0 | XP1 | XP2 | XP3 | XP4 | AP1 | AP2 |
| TN | 37 | 46 | 0 | 64 | 63 | 0 | 67 |
| FP | 29 | 13 | 42 | 1 | 2 | 22 | 0 |
| FN | 3 | 6 | 0 | 14 | 10 | 0 | 12 |
| TP | 30 | 27 | 31 | 19 | 23 | 28 | 19 |
| Uncertain | 1 | 8 | 27 | 2 | 2 | 50 | 2 |

Abbreviations: TN: true negative; FP: false positive; FN: false negative; TP: true positive